\documentclass[10pt]{cai}

\graphicspath{{figs/}}

\begin{document}

\volumeheader{36}{0}
\begin{center}

  \title{GUILGET: GUI Layout GEneration with Transformer}
  \maketitle

  \thispagestyle{empty}

  
  \begin{tabular}{cc}
    Andrey Sobolevsky\upstairs{\affilone,*}, Guillaume-Alexandre Bilodeau\upstairs{\affilone}, Jinghui Cheng\upstairs{\affilone}, Jin L.C. Guo\upstairs{\affiltwo}
   \\[0.25ex]
   {\small \upstairs{\affilone} Polytechnique Montréal} \\
   {\small \upstairs{\affiltwo} McGill University} \\
  \end{tabular}
  
  \emails{
    \upstairs{*}andrey.sobolevsky@polymtl.ca 
    }
  \vspace*{0.2in}
\end{center}

\begin{abstract}
Sketching out Graphical User Interface (GUI) layout is part of the pipeline of designing a GUI and a crucial task for the success of a software application. Arranging all components inside a GUI layout manually is a time-consuming task. In order to assist designers, we developed a method named GUILGET to automatically generate GUI layouts from positional constraints represented as GUI arrangement graphs (GUI-AGs). The goal is to support the initial step of GUI design by producing realistic and diverse GUI layouts. The existing image layout generation techniques often cannot incorporate GUI design constraints. Thus, GUILGET needs to adapt existing techniques to generate GUI layouts that obey to constraints specific to GUI designs. GUILGET is based on transformers in order to capture the semantic in relationships between elements from GUI-AG. Moreover, the model learns constraints through the minimization of losses responsible for placing each component inside its parent layout, for not letting components overlap if they are inside the same parent, and for component alignment. Our experiments, which are conducted on the CLAY dataset, reveal that our model has the best understanding of relationships from GUI-AG and has the best performances in most of evaluation metrics. Therefore, our work contributes to improved GUI layout generation by proposing  a novel method that effectively accounts for the constraints on GUI elements and paves the road for a more efficient GUI design pipeline.
\end{abstract}

\begin{keywords}{Keywords:}
Graphical User Interface, GUI arrangement graphs, deep learning, transformer, generative model, GUI layout
\end{keywords}
\copyrightnotice

\section{Introduction}
\label{intro}
\begin{figure}[htbp]
    \centering
    \includegraphics[scale=0.5]{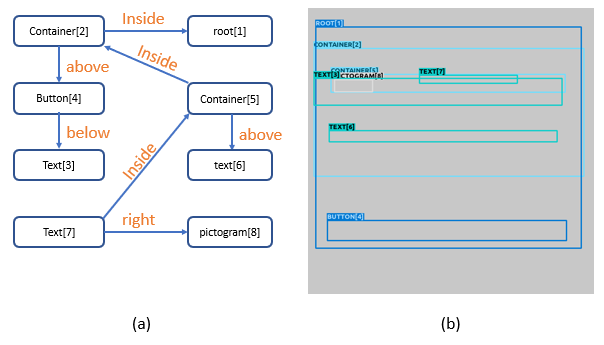} \\
    \caption{GUI layout generation goal is to generate a realistic GUI layout (b) from a given GUI-AG (a).}
    \label{fig:intro}
\end{figure}

The design of Graphical User Interface (GUI) is an important aspect that affects the success of many software applications. The first step for the GUI designers is often sketching out the interface layout with wireframes, based on design constraints such as users' needs and software requirements~\cite{hartson2018ux}. These GUI layouts define the visual arrangement of elements in a user interface, such as buttons, text fields, and containers. Creating these layouts and variations of them manually, however, can be time-consuming. In this paper, we explore an automated technique that can support designers creating GUI layouts. In our approach, we capture the design constraints that the designers have to consider through \textit{GUI arrangement graphs} (GUI-AGs) and generate graphical user interface layouts from those constraints. GUI-AGs specify elements required in the UI design and the relationships among them, as illustrated in \autoref{fig:intro}. We use GUI-AGs because they can be used to describe the requirements with an explicit definition of components that are part of the screen and the definition of the visual relations between those components. These graph models offer the flexibility to automatically create layout variations by modifying relations in the graph. 


There are several technical challenges for achieving automatically generatation of GUI layouts from GUI-AGs. One challenge is to accurately capture the logical and semantic relationships between GUI elements, such as hierarchical structures and functional dependencies. Another challenge is to generate visually appealing and functional layouts that adhere to design principles and constraints \cite{tidwell2010designing}. In addition, the generation process should be efficient to be used in design workflow. To address these challenges, we propose a transformer-based approach for generating GUI layouts from GUI-AGs.  Transformer networks have recently achieved state-of-the-art results in a wide range of natural language processing and computer vision tasks. They are particularly well-suited for generating GUI layouts from GUI-AGs, as they can effectively capture the dependencies between elements in the GUI-AG and generate GUI layouts that reflect these dependencies. Our transformer-based model takes as input a series of tokens that express the GUI-AG relationships; it then outputs a realistic GUI layout. We demonstrate the effectiveness of our approach through a series of experiments on real-world datasets \cite{li2022learning, deka2017rico} using metrics specific to this task. Results indicated that our approach produces the most relevant GUI layouts in regard to specified requirements.

Our contributions can be summarized with the following:
\begin{itemize}
    \item We propose a new transformer-based method to generate GUI layouts from GUI-AGs that takes into account the GUI design constraints. Our experiment demonstrates that it better captures the intended GUI layout compared to previously proposed methods \cite{johnson2018image, yang2021layouttransformer};
    \item We introduce new loss functions and metrics for quantitative measurement of the quality of generated GUI layouts; Those metrics can be used for future work on similar tasks.
    \item To encourage reproduction or replication of our study, the source code of our experiment is publicly available with a pre-trained model at \url{https://github.com/dysoxor/GUILGET}.  
\end{itemize}

\section{Related Work}


GUI generation is an emerging but yet under-explored computer vision application, especially for GUI generation based on GUI-AGs. First studies on generating GUI designs automatically adopt generative adversarial network, such as GUIGAN \cite{zhao2021guigan} and GANSpiration \cite{mozaffari2022ganspiration}. However, those approaches respectively do not generate new components style and do not consider any specification of the design -- what components should be included or the relation between them. GUIGAN is based on a sequence of subtrees, which are hierarchical tree structures made of components, which is used as input and produces GUI by reusing different components based on their style without having the ability to produce new components. On the other hand, GANSpiration produces new design examples from existing screenshots or random vectors from the latent space representing the screenshots. To the best of our knowledge, no existing GUI generation methods explicitly consider the specifications about the content, including the relationship between components. However, there are several works focusing on image and layout generation for more general domains that can be inspirational for GUI generation \cite{li2019pastegan, zhao2022high}. For example, PasteGAN \cite{li2019pastegan} generates images from scene graphs and image crops by taking into account the semantic relationships between objects and their visual appearance. Another method, proposed by Zhao et al. \cite{zhao2022high}, uses a transformer to generate image tokens from a given scene graph and then decodes those tokens into a plain image using a VQGAN decoder \cite{esser2021taming}. In comparison, our work aims to produce first GUI layouts, which is usually the first step in the GUI design process, rather than directly generating the GUI design. Our work can be used as a part of a GUI design generation pipeline.

Our concept of GUI-AG is inspired by scene graph, a widely used structure in computer graphics. The task of generating layouts from scene graphs is studied in computer graphic, but also requires knowledge from natural language processing (NLP) to process the textual input and understand the interactions between each element of the scene graph. Scene graphs are commonly used to represent the structure and relationships between graphical elements in a scene, and layout generation aims to determine the position and size of these elements in the final layout.

There have been several approaches proposed for generating layouts from scene graphs in the literature. For example, SG2IM by Johnson et al. \cite{johnson2018image} adopted a graph convolutional network (GCN) to process the scene graph. This approach, however, is unable to capture semantic equivalence in graphs. The work of Herzig et al. \cite{herzig2020learning} tackle this issue by learning canonical graph representations from the input scene graph, and it allows to represent more complex scene graphs. Another way of processing scene graphs is by using transformer-based model such as the ones presented in \cite{gupta2021layouttransformer,yang2021layouttransformer}. Transformers have shown great performances in processing textual components and learning relations between them. The method presented by Gupta et al. \cite{gupta2021layouttransformer} uses self-attention to learn contextual relationships between layout element. In practice, it takes layout elements as input and produces the next layout elements as output. So it has no constraints and the model produces random new layouts. On the other hand, in the work LayoutTransformer by Yang et al. \cite{yang2021layouttransformer}, the transformer-based model takes as input constraints in the form of scene graphs and predicts the layout based on relationships.

Many works have been done to generate layouts with different type of input from scene graphs. LayoutVAE \cite{jyothi2019layoutvae}, a variational autoencoder (VAE)-based framework for generating stochastic scene layouts, is a model that takes as input a set of labels representing the entities that must appear in the layout. It does not take into account any positional constraints, neither the amount of each label as with scene graphs. The Neural Design Network (NDN) \cite{lee2020neural} combines a GCN with a conditional VAE to generate layouts based on constraints given as input. In NDN, the scene graph is generated based on the amount of components there are and some desired positional constraints. Another work uses a generative adversarial network (GAN) \cite{li2019layoutgan} to generate layouts from an input of randomly-placed 2D elements. It uses self-attention modules to refine their labels and geometric parameters jointly to produce a layout.

In contrast to previous works, we propose a method that takes into account the specifications of a UI design using GUI-AGs that are considering the amount of each UI component and component position relative to other components.

\section{Method}
\begin{figure}[htbp]
    \centering
    \includegraphics[scale=0.33]{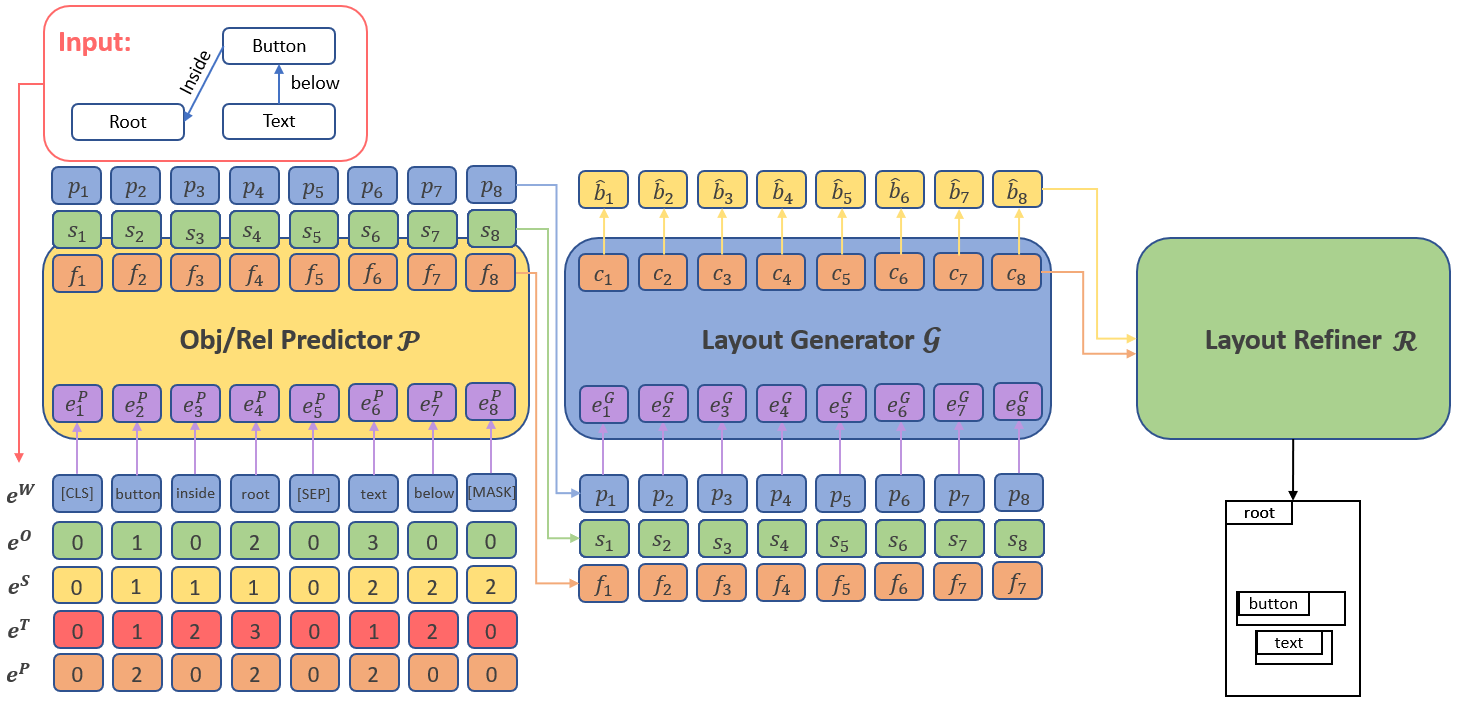} \\
    \caption{Architecture of our model with three main components. \textbf{Obj/Rel Predictor $\mathcal{P}$} takes as input an embedding $e^P$ which is a concatenation of several embeddings that describe different information about the given GUI-AG, and it produces contextual features $f$. \textbf{Layout Generator $\mathcal{G}$} takes as input contextual features $f$, predicted sizes $s$ and predicted positions $p$ to translate it into a layout-aware representation $c$ and bounding boxes $b$. \textbf{Layout Refiner $\mathcal{R}$} uses co-attention module with predicted bounding boxes $b$ and layout-aware representation $c$ to improve the layout.}
    \label{fig:architecture}
\end{figure}
Our work is based on the use of transformers \cite{vaswani2017attention} that have been shown to be state-of-the-art in multiple natural language processing tasks. A transformer uses the concept of self-attention, which allows to give different weights depending on the part of the input to make predictions. GUI-AGs can be then viewed as a natural language input to the transformer since it is a logical sequence of relationships.  

Our proposed method is based on the LayoutTransformer (LT-net) \cite{yang2021layouttransformer}, a transformer-based model that aims to generate diverse layouts from scene graphs of images. It consists of (1) an object/relation predictor $\mathcal{P}$ that encodes the input scene graph into contextual features $f$ using transformer encoder layers, (2) a layout generator $\mathcal{G}$ that, from contextual features $f$, generates bounding boxes $b$ with distributions matching a learned Gaussian distribution model and layout-aware representations $c$ with transformer decoder layers, and (3) a layout refiner $\mathcal{R}$ made of a co-attention module. Our architecture is presented in \autoref{fig:architecture}. Compared to LT-net, we introduced several improvements for GUI layout generation that will be presented in the following.

\subsection{Building a GUI arrangement graph}
\begin{figure}[htbp]
    \centering
    \includegraphics[scale=0.33]{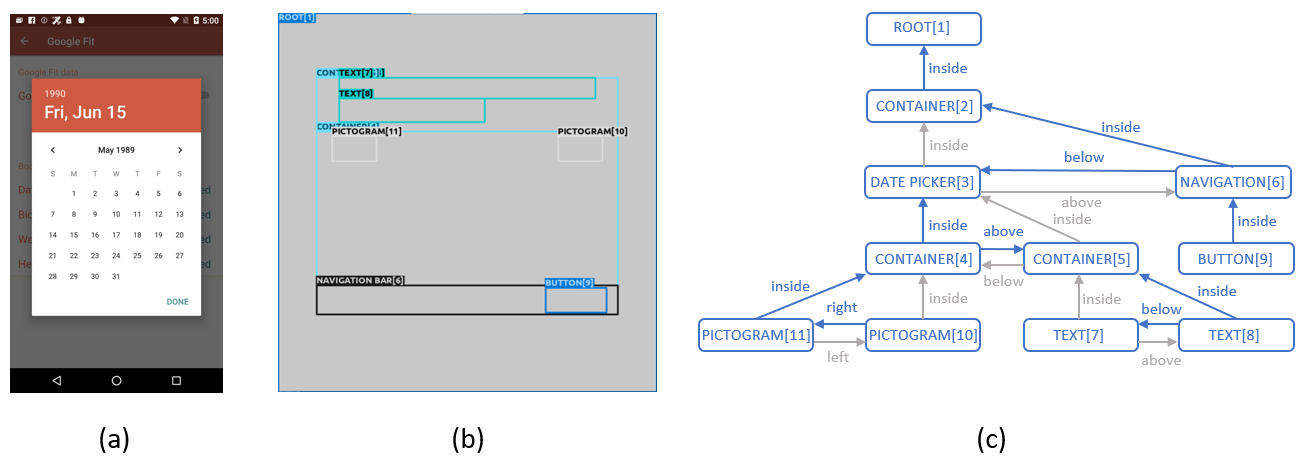} 
    \caption{Heuristic process of modeling GUI-AG (c) based on given layout (b) and associated screenshot (a). All nodes represent components and all arrows represent possible relationships. We will keep only one \textit{inside} relation among all children from a component (in blue) randomly, while the others are not used as input (in gray). The other relations are only possible between components that are inside the same parent and we keep only one relation between them.}
    \label{fig:sgmodeling}
\end{figure}

GUIs are made of two types of components: (1) widgets (e.g. button, pictogram, text), which are leaf nodes in a GUI-AG representation and do not contain any other component, and (2) spatial layouts (e.g. container, list item, toolbar), which are intermediate nodes that allow to organize the structure of widgets \cite{zhao2021guigan}. This tree structure does not exist in images and it changes the way GUI-AGs should be modeled and processed compared to scene graphs for images. In contrast to our work, most of current works that are done in layout generation for GUI ignore the tree structure and only consider widgets to remove the complexity of organizing those widgets inside layouts. 

GUI-AGs are directed graphs made of relationships that are triplets of \textit{subject-predicate-object} \cite{johnson2018image}. A GUI does not require as much variety of relations and objects as images. However, it has complexity since GUI layouts have more rules and principles to be learned than in a layout from an image \cite{tidwell2010designing}. To train a neural network model, GUI-AGs from ground truth layouts are required. We define five types of possible predicate specifying the relationship among GUI components: \textit{left, right, top, bottom} and \textit{inside}. We build a GUI-AG from a layout by parsing a layout description. First, to reduce the size of the GUI-AG in order to have a smaller input in the transformer, we keep randomly only one \textit{inside} relation within a group of components that are children to the same parent. This step is beneficial due to memory limitations when the input is too large. To illustrate this step, in \autoref{fig:sgmodeling} we notice that \textit{CONTAINER[2]} has two children, \textit{DATE PICKER[3]} and \textit{NAVIGATION[6]}, but only one \textit{inside} relationship is kept as input to the model and the other one is removed. This processing method does not lose information since all components that do not have an \textit{inside} relationship are considered at the same level and are implicitly inside the parent component. We also randomly choose a sequence of components inside the layout and determine relationship between each pair of components within the sequence. More precisely, as shown in \autoref{fig:sgmodeling}, \textit{DATE PICKER[3]} and \textit{NAVIGATION[6]} are inside the same parent; in this situation, we randomly choose the sequence in the set of children \textit{[NAVIGATION[6], DATE PICKER[3]]} and then add the relation between each component of the sequence to form triplets (\textit{[NAVIGATION[6], below, DATE PICKER[3]]} in this case). By doing so, we get a simplified input that captures most information from the graph, as we will see in \autoref{GUI-AG embedding}. It is to note however, that we may miss some relationships. For example, if there are three components (a,b,c) and we keep only two relationships (a-b, b-c), the relation between a-c can be uncertain in some cases. Finally, in GUI-AGs, the relationships are reversible; e.g., if in the layout from \autoref{fig:sgmodeling} the \textit{NAVIGATION[6]} is below \textit{DATE PICKER[3]}, the reverse, \textit{DATE PICKER[3]} is above \textit{NAVIGATION[6]}, must also be true. Hence, we can keep only one relationship between them.

\subsection{Object/Relation Predictor}
\label{GUI-AG embedding}

To construct the input for our transformer model, we first convert the GUI-AG into a sequence of relationship triplets $s_i$. We refer to this sequence of relationships as $S = \{s_1, s_2,... s_T\}$ where $T$ is the number of relationships. Relationships are separated by a special token \textit{SEP}, and a special token \textit{CLS} is used at the beginning of the entire sequence $S$.

Instead of using directly the sequence $S$ as input in the Object/Relation Predictor, we embed $s_{1:T}$ into $e_{1:T}^P$, which allows to take into consideration different features: word embedding $e_{1:T}^w$ that allows to identify the class of the object (e.g. "button" or "container") or the relation (e.g. "inside" or "right), object ID embedding $e_{1:T}^o$ to differentiate instances of the same object, relationship ID embedding $e_{1:T}^s$ to separate each relationship, type of word embedding $e_{1:T}^t$ to identify parts of a relationship (\textit{subject = 1, predicate = 2, object = 3)} for example the part of the the relationship "button inside container" are "subject predicate object" and instead of writting the plain text, we rather associate an ID to each part of the relationship to differentiate them, and parent ID embedding $e_{1:T}^p$, which is a feature that allows for each of the component to know its parent. Those features are concatenated to form the input embedding for the object/relation predictor. It is given by

\begin{equation}
    \label{eq:embedding}
    e_{1:T}^P = [e_{1:T}^w \bigoplus e_{1:T}^o \bigoplus e_{1:T}^s \bigoplus e_{1:T}^t \bigoplus e_{1:T}^p].
\end{equation}

The object/relation predictor learns to produce three different outputs: (1) the contextualized feature vectors $f_{1:T}$, (2) the size vectors $s_{1:T}$, and (3) the position vectors $p_{1:T}$. The contextualized feature vectors $f_{1:T}$ describe objects, relations and their context with features from the input. In order to capture conceptually diverse embedding and exploiting the co-occurrence among objects, predicates and parents, we follow the technique used in BERT \cite{devlin2018bert} and mask randomly words from the input that must be predicted by the object/relation predictor. The size vectors $s_{1:T}$ are predictions of the bounding boxes size for each object made by the object/relation predictor. It is used as indicator later in the GUI layout generator to generate final bounding boxes size. The position vectors $p_{1:T}$ are predictions of bounding boxes position for each object.

Finally, in order to compute the objective function to train this part of the network, we need to predict $\hat{e}^P_{1:T}$ from the features $f_{1:T}$ using a single linear layer. Indeed, since there is not ground truth for $f_{1:T}$, we predict $\hat{e}^P_{1:T}$ from $f_{1:T}$ and match it with $e^P_{1:T}$ to minimize the reconstruction error. The objective function for training the module is composed of cross-entropy losses $\mathcal{L}_{predSem}$, given by
\begin{equation}
    \mathcal{L}_{predSem} = CrossEntropy(e^P_t,\hat{e}^P_t),
\end{equation}
for the matching word, object ID, type of word, parent ID which are all extracted from the input GUI-AG, and regression losses given by 
\begin{equation}
    \mathcal{L}_{predBox} = Regression(s_t,\hat{s}_t) + Regression(p_t,\hat{p}_t),
\end{equation}
which are computed on predicted positions $\hat{p}_{1:T}$ and sizes $\hat{s}_{1:T}$ with their corresponding ground truth positions and sizes. The total loss of the predictor $\mathcal{L}_{pred}$ is a combination of the two losses and given by
\begin{equation}
    \mathcal{L}_{pred} = \mathcal{L}_{predSem} + \mathcal{L}_{predBox}.
\end{equation}

\subsection{Layout generator }
The goal of the layout generator module is to produce layout-aware representations $c_{1:T}$ and bounding boxes $\hat{b}_{1:T}$. This module is made of transformer decoder layers and interprets jointly and sequentially contextual features $f_{1:T}$, predicted bounding box sizes $s_{1:T}$ and predicted bounding box positions $p_{1:T}$ that are computed by the object/relation predictor module. The three given inputs are concatenated and expressed as $e^G_{1:T}$. After that it is translated into diverse bounding box output $\hat{b}_{1:T}$. A bounding box is described by its top-left corner position in a 2D Cartesian coordinate system and its size in terms of width and height, i.e., $\hat{b}_t = (x_t, y_t, w_t, h_t)$, and there is a bounding box produced for each subject, predicate and object. The bounding box of the predicate is the difference between the position of the object and the one of the subject, i.e., $\hat{b}_t = (x_{t+1} - x_{t-1}, y_{t+1} - y_{t-1})$.

To produce sequentially the bounding boxes $\hat{b}_t$, the features from the input $e^G_t$ are also concatenated with the previously produced bounding box $\hat{b}_{t-1}$. This input is not directly translated into the bounding box but to a layout-aware representation $c_t$ that is used to model a distribution in order to use \textit{Gaussian Mixture Models (GMM)} \cite{reynolds2009gaussian}. We use this instead of directly predicting the bounding box from layout-aware representation in order to have a generative ability. Given a bounding box  distribution, the bounding box $\hat{b}_t$ will be sampled from the posterior distribution $p_{\theta_t}(\hat{b}_t|c_t)$ knowing $c_t$. It can be described as follows:

\begin{equation}
    p_{\theta_t}(\hat{b}_t|c_t) = \sum^K_{i=1} \pi_i\mathcal{N}(\hat{b}_t;\theta_{t,i}),
\end{equation}

\noindent where $i$ indicates the $i$-th distribution out of $K$ multivariate normal distributions, $\theta_{t,i}$ are the parameters of each distribution defined by $(\mu_{t,i}^x, \mu_{t,i}^y, \sigma_{t,i}^x, \sigma_{t,i}^y, \rho_{t,i}^{xy})$ where $\mu$, $\sigma$ and $\rho$ denote respectively the mean, standard deviation and the correlation coefficient, $\pi_i$ is a magnitude factor, and $\mathcal{N}$ is the multivariate normal distribution.

To define the objective function of the generator $\mathcal{L}_{gen}$, we start by defining the box reconstruction loss $\mathcal{L}_{box}$, which maximizes the log-likelihood of the generated GMM to fit the training data where the ground-truth bounding boxes are denoted as $b_t=(x_t, y_t, w_t, h_t)$.

\begin{equation}
    \mathcal{L}_{box} = -\frac{1}{K}\log ( \sum^K_{i=1} \pi_i\mathcal{N}(b_t;\theta_{t,i})).
\end{equation}

To avoid the over-fitting with this loss function, the GMM distributions are fitted to a multivariate normal distribution $Q$ using a Kullback-Leibler (KL) divergence loss:

\begin{equation}
    \mathcal{L}_{KL} = \sum^K_{i=1} D_{KL}(P_i||Q_i).
\end{equation}

Finally, a relation consistency loss $\mathcal{L}_{rel}$ is also used since the two previous losses focuses only on the bounding boxes. It is given by:

\begin{equation}
    \mathcal{L}_{rel} = \frac{1}{N} \sum (\Delta \hat{b}_t - \hat{b}^{rel}_t)^2,
\end{equation}
where $N$ denotes the number of relationships in S. It calculates the Mean Square Error (MSE) between the box disparity of the relation we get, i.e. $\hat{b}_t^{rel}$ which is the predicted bounding box for the predicate, and the corresponding box disparity we calculate from the object and the subject $\Delta \hat{b}_t = (x_{t+1} - x_{t-1}, y_{t+1} - y_{t-1})$. The layout generator is trained by minimizing the weighted sum of losses using

\begin{equation}
    \mathcal{L}_{gen} = \lambda_{box} \mathcal{L}_{box} + \lambda_{KL} \mathcal{L}_{KL} + \lambda_{rel} \mathcal{L}_{rel},
\end{equation}

\noindent where  $\lambda_{box}$, $\lambda_{KL}$ and $\lambda_{rel}$ are weighting factors for each corresponding loss.

\subsection{Layout refiner}

Since the bounding boxes are generated sequentially, they require refinement in the layout in order to consider the semantic $c_{1:T}$ and the bounding box $\hat{b}_{1:T}$. This is done in the layout refiner using the Visual-Textual Co-Attention (VT-CAtt) \cite{yang2021layouttransformer}, which predicts the residual $\Delta \hat{b}_{1:T}$ for updating the bounding boxes. The objective function of this module $\mathcal{L}_{ref}$ is defined with multiple losses. The first one is a regression loss $\mathcal{L}_{reg}$ between predicted bounding boxes $b'_{1:T}$ by the layout refiner and the ground truth bounding boxes $b_{1:T}$. Another new and task-specific loss that we implemented is the overlap between children loss $\mathcal{L}_{CC}$, which aims to minimize the overlap of components that share the same parent in the GUI interface. This is specific to design principles in GUI since we do not want the components to overlap because it will hide some components on the final interface. This loss is given by 

\begin{equation}
    \mathcal{L}_{CC} = \frac{C_1 \cap C_2}{\min(C_1, C_2)}, 
    \label{eq:lcc}
\end{equation}

\noindent where $C$ designate the area of a children. Another principle to follow is that a children must be inside its parent. To enforce this principle, we define the overlap between children and parent loss $\mathcal{L}_{CP}$ and we express it as

\begin{equation}
    \mathcal{L}_{CP} = 1 - \frac{C \cap P}{C}, 
    \label{eq:lcp}
\end{equation}

\noindent where $C$ is the area of the children and $P$ is the area of its parent that is defined in the input GUI-AG. The objective function on which the layout refiner is trained is a weighted sum of those losses, that is

\begin{equation}
    \mathcal{L}_{ref} = \lambda_{reg} \mathcal{L}_{reg} + \lambda_{CC} \mathcal{L}_{CC} + \lambda_{CP} \mathcal{L}_{CP}.
\end{equation}

\noindent where  $\lambda_{reg}$, $\lambda_{CC}$ and $\lambda_{CP}$ are weighting factors for each corresponding loss.

\section{Experiments}
In our experiment, we validated our proposed method by generating layouts from GUI-AGs. We aim to evaluate how close the generated layouts are to the ground truth layouts based on the graphs associated to them. 

\subsection{Dataset}
We tested our method on the \textbf{CLAY dataset} \cite{li2022learning}, which is a UI design dataset. UI layouts in RICO dataset \cite{deka2017rico} are often noisy and have visual mismatches hence CLAY is a dataset that improves RICO by denoising UI layouts. It contains 59,555 human-annotated screen layouts, based on screenshots and layouts from RICO. A total of 24 component categories (e.g. Image, button, text) are available in layouts and 5 predicate categories (above, below, right, left, inside) are considered in GUI-AGs. 

By observing data from the CLAY dataset, we decided to remove several irrelevant GUIs and their layouts. Firstly, we removed GUIs that contain two or less types of components. Those screenshots are usually not representing an application GUI but rather GUIs with, for instance, full screen image and video screenshot that do not contain useful information for our task, as there is a lack of component and interactions (predicates) between components. Then, we also removed screens that contain only a navigation bar or popup for example, for the same reason. Also usually in the dataset there are several screenshots associated to the same application but some contain only the navigation bar and others have the navigation bar and also some content. So we removed the former to avoid overfitting. In practice, we achieved that by removing GUIs in which components cover less than 25\% of the total area of the screen.


\subsection{Evaluation metrics}
\label{metrics}
To evaluate the generated layouts with respect to the ground truth, we used the following metrics. 

\textbf{CP Inclusion (CPI)} is a metric that captures the overlapping of children with its parent. The metric is computed as $1-\mathcal{L}_{CP}$ (see \autoref{eq:lcp}). Hence, the goal of this metric is to indicate if the generated GUI layouts tend to satisfy the UI design principle that states that children must be fully inside its parent as we see in \autoref{tab:comparison} with ground truth data.

\textbf{CC Separation (CCS)} is another metric that aims to evaluate if the UI design principles tend to be satisfied. It is computed by $1-\mathcal{L}_{CC}$ (see \autoref{eq:lcc}), so the metric measure the ratio of components that does not overlap between each other and in the same time share a common parent.

\textbf{Alignment} \cite{lee2020neural} metric evaluates an important design principle that is components must be either in center alignment or in edge alignment (i.e. left-, right-, bottom- or top-aligned). We computed alignment with the following:

\begin{equation}
    1 - \frac{1}{N_C} \sum_d \sum_i \min_{j,i\neq j}\{\min (l(c_i^d, c_j^d), m(c_i^d, c_j^d), r(c_i^d, c_j^d), t(c_i^d, c_j^d), v(c_i^d, c_j^d), b(c_i^d, c_j^d))\},
\end{equation}

\noindent where $N_C$ is the number of components, $c_k^d$ is the $k_{th}$ component of the $d_{th}$ layout and $l$, $m$, $r$, $t$, $v$ and $b$ are alignment functions where the distance between the left, horizontal center, right, top, vertical center and bottom are measured, respectively.

\textbf{W bbox} is the similarity between bounding box properties $(x_{left}, y_{top}, w, h)$ distribution of the generated GUI layouts and the ground truth GUI layouts. This is computed using Wasserstein distance and inverting it to be a similarity, between 0 and 1, by subtracting the maximum possible distance by the actual distance and normalizing the value. It is a way to measure if the generated bounding boxes are as diversified as in the ground truth data.


\textbf{GUI-AG Correctness (GUI-AGC)} computes the average number of correct relationships that appears in the generated GUI layout. In practice, we compare the input GUI-AG with the corresponding generated GUI layout and count the number of satisfied relationships divided by the total number of relationships.

\subsection{Quantitative results}
\label{quantitative}

\begin{table}[htbp]

\begin{center}
\begin{tabular}{|c|c|c|c|c|c|c|}
\hline
\textbf{}&\multicolumn{5}{|c|}{\textbf{Metrics}} \\
\cline{2-6} 
\textbf{} & \textbf{CPI $\uparrow$}& \textbf{CCS $\uparrow$}& \textbf{Alignment $\uparrow$}& \textbf{W bbox $\uparrow$}& \textbf{GUI-AGC $\uparrow$} \\
\hline
GT data& 1.0 & 0.987 & 1.0 & - & -\\
\hline
SG2IM \cite{johnson2018image}& 0.191 & \textbf{0.974} & 0.997 & 0.81 & 0.369\\
LayoutTransformer \cite{yang2021layouttransformer}& 0.392 $\pm$ 0.001 & 0.805 $\pm$ 0.002 & 0.998 $\pm$ 2E-5 & \textbf{0.834} $\pm$ 1E-4 & 0.797 $\pm$ 0.001\\
\hline
GUILGET (ours)& \textbf{0.592} $\pm 0.001$ & $0.623 \pm 0.002$ & \textbf{0.9983} $\pm$ 2E-5 & 0.811 $\pm$ 8E-5 & \textbf{0.868} $\pm 0.001$\\
\hline
\end{tabular}
\end{center}
\caption{Quantitative evaluation on CLAY dataset \cite{li2022learning}. 3000 layouts are generated with each method and are compared using metrics presented in \autoref{metrics}. For SG2IM \cite{johnson2018image} as well as for LayoutTransformer \cite{yang2021layouttransformer}, we only consider the parts of the architectures responsible to generate the layout.}
\label{tab:comparison}
\end{table}

\autoref{tab:comparison} gives the results of our method compared to SG2IM \cite{johnson2018image} and to LayoutTransformer \cite{yang2021layouttransformer}. As we can observe, there are several metrics where our model gives the best results but not with all metrics. If we compare our model to SG2IM, we can see that in terms of \textit{CCS} we do not get as good results. However, we notice that \textit{CPI} and \textit{GUI-AGC} are the worst for SG2IM. In particular, \textit{GUI-AGC} shows that only $36.9\%$ of the relations given as input to SG2IM are satisfied in the output, which means that the design constraints are not met with this method. Moreover, we can see with this model that the \textit{CCS} is high while \textit{CPI} is low, which means that SG2IM does not organize components inside the layout but uses the whole screen, as it is also shown in the results from \autoref{fig:qualitative}, which makes it easier to get a high \textit{CCS} since it does not learn the \textit{inside} constraint. It is easy not to have overlap between child components in that case. In other words, it is not meaningful to have a high \textit{CCS} if \textit{CPI} and \textit{GUI-AGC} are not also high. The LayoutTransformer model has more understanding of predicates but it is still worse than with our model. We can also see that there is a negative correlation between \textit{CCS} and \textit{CPI} -- if the model learns to place components inside its parent, there are more possible overlaps between components inside a layout. We want both of these metrics to be similarly high to respect both of those GUI design constraints as we can observe it in the GT data, where all metrics related to GUI design constraint are close to 1. \textit{W bbox} metric is similar for all of the models which is understandable since our model and LayoutTransformer model generate bounding boxes based on distribution of bounding boxes from the training GUI layouts. On the other hand, SG2IM is not a generative model and aims to predict bounding boxes which leads to learn the most common sizes and positions for different types of component.

\subsubsection{Influence of UI category}

\textbf{\begin{figure}[htbp]
    \centering
    \includegraphics[scale=0.33]{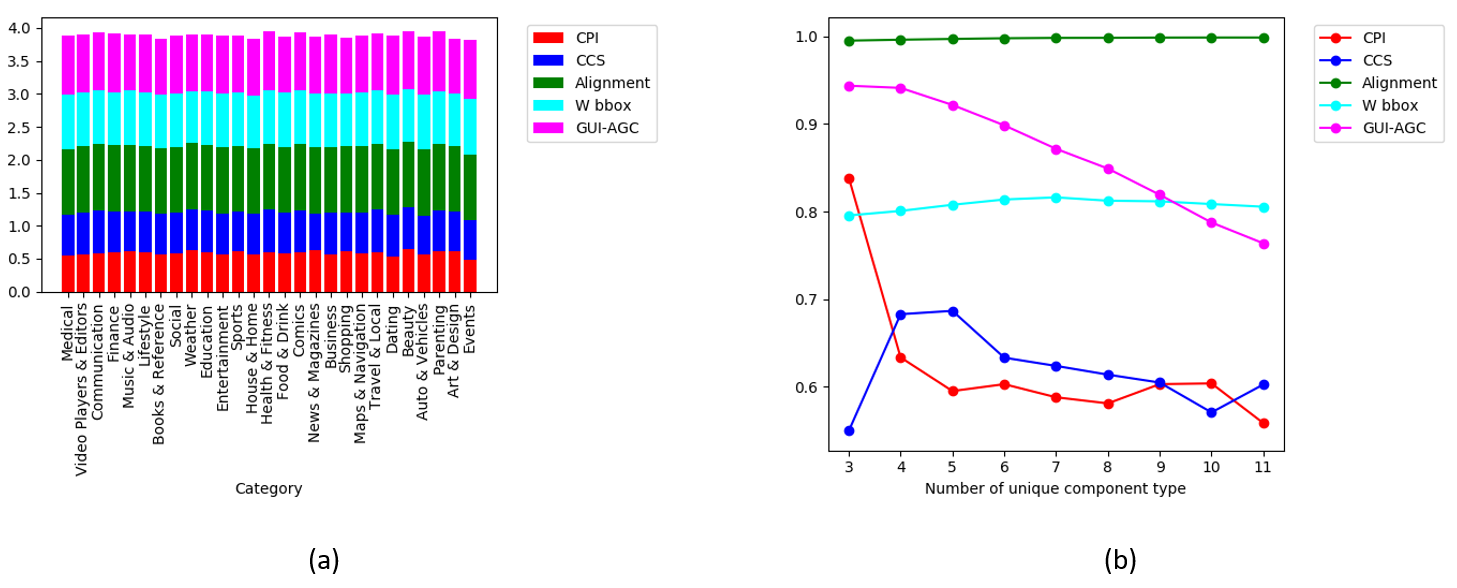} \\
    \caption{Quantitative evaluation on different screen categories (a) from CLAY dataset \cite{li2022learning}. Evaluation metrics are applied on all 27 screen categories separately. (b) shows the influence of number of unique type components on evaluation metrics.}
    \label{fig:influence}
\end{figure}}

In order to see if the screen category has an impact on the performance, we conducted an experiment where we compute each evaluation metric for each category separately. \autoref{fig:influence} (a) summarizes the results. Overall, our model yields similar performances among different app categories. This is a conclusive result which shows that our model is not biased toward certain types of screen categories and is able to produce equally good GUI layout for any of the category.

\subsubsection{Influence of UI complexity}


Similarly as with the previous experiment, we want to understand the influence of the UI complexity (indicated by the number of unique component types in the UI~\cite{mozaffari2022ganspiration}) on the performance. \autoref{fig:influence} (b) shows that with smaller number of unique component types most of the evaluation metrics are better; in other words, our model achieved better performances when the UI is less complex. Particularly, the \textit{GUI-AGC} metric is inversely proportional to the number of unique component types in the UI. For both \textit{CCS} and \textit{CPI} metrics, these results are expected due to the fact that with more diverse components that have various sizes and position standards, it becomes harder to organize all elements inside spatial layouts. There are however two metrics performing equally well over all number of unique component types, which are the \textit{Alignment} and \textit{W bbox} metrics. This shows that our model succeeds to always align components whether the complexity is low or high, and generated bounding boxes have almost the same similarity in distribution with the ground truth distribution.

\subsection{Qualitative results}

\begin{figure}[htbp]
    \centering
    \includegraphics[scale=0.33]{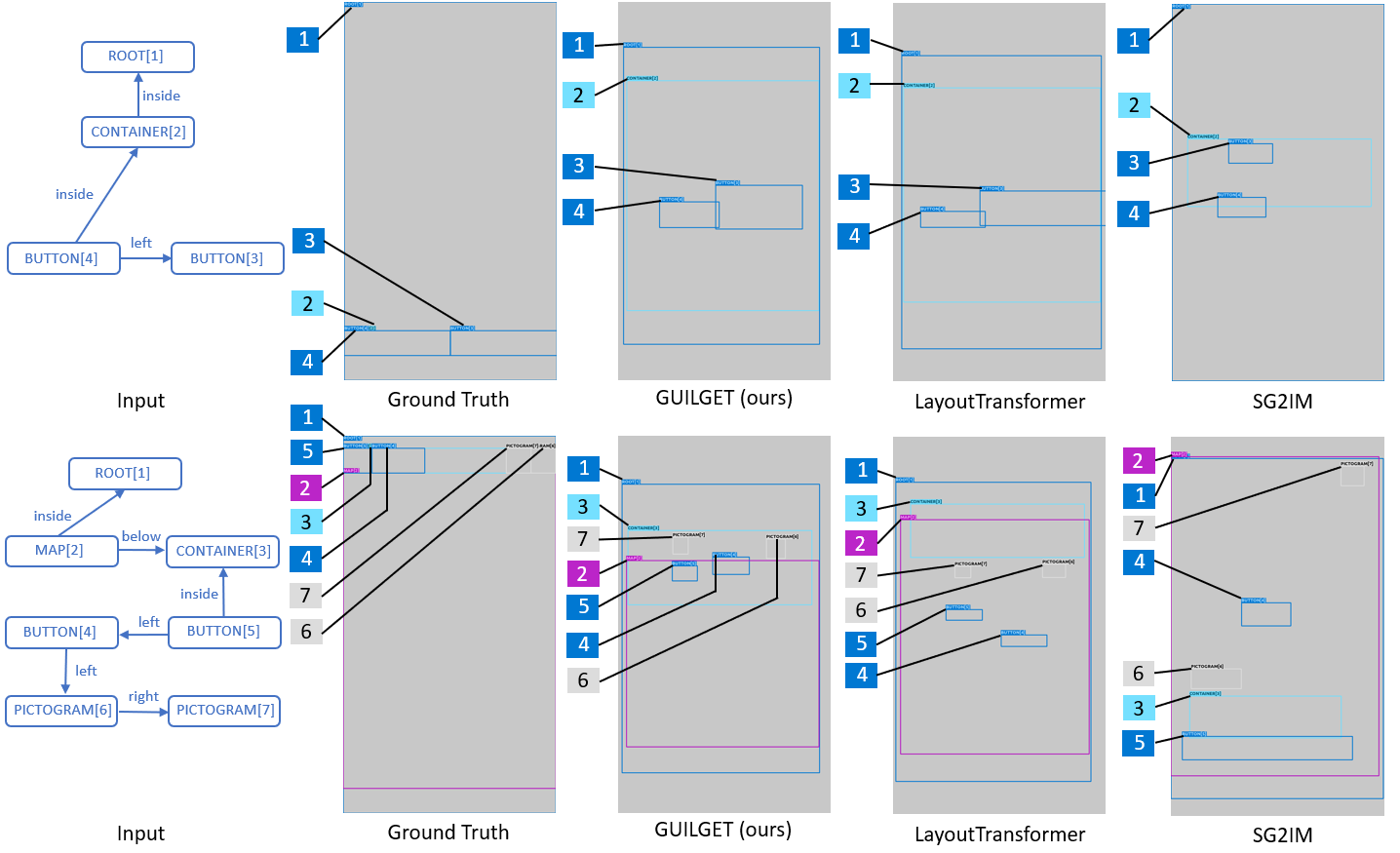} \\
    \caption{Qualitative comparison between our model, LayoutTransformer and SG2IM. The same input is given for the three models. The input from the first row has a low complexity with 3 unique component types while the second input has a larger complexity with 5 unique component types.}
    \label{fig:qualitative}
\end{figure}
\autoref{fig:qualitative} shows two examples produced by our model, LayoutTransformer, and SG2IM. Examples were chosen manually based on number of unique component types inside the GUI layout. We show results for low complexity (3-4 unique component types) and medium complexity (5-7 unique component types). We do not show GUI With large complexity (8 or more unique component types) as it is harder to analyze visually because of the larger number of components. The results of \autoref{fig:qualitative} are aligned with the quantitative results from \autoref{quantitative}. Indeed, we can see that SG2IM has a poor semantic understanding of relations for both cases in \autoref{fig:qualitative}; it also struggles to place components inside their parent as we can observe in the second row from \autoref{fig:qualitative} where all components that are supposed to be inside the \textit{CONTAINER[3]} are not and the container which is supposed to be below the \textit{MAP[2]} is actually entirely inside it instead. We can note however that the sizes of bounding boxes and their alignments are realistic. The LayoutTransformer shows a better understanding of relations but is not able in the second case to place components inside its parent as exemplified by \textit{BUTTON[4]} and \textit{BUTTON[5]} that are outside the \textit{CONTAINER[3]} in the second row from \autoref{fig:qualitative}. In contrast, our model respects all the given constraints and placed correctly buttons inside the container. Also, GUILGET generates plausible bounding boxes even though the generated layouts are not aligned in the way it is in the ground truth. However, information from GUI-AG is not complete enough to reproduce the same alignment. Adding global positioning constraints on components to the GUI-AG could be an interesting avenue to investigate. 

\section{Conclusion}
This work propose a transformer-based model that generates a GUI layout from a given GUI-AG. Our approach is the state-of-the-art in quantitative performance across several metrics and in visual quality. We saw that using attention provides a higher performance than using graph convolution network in capturing semantic of the GUI-AG. The new components from our model compared to LayoutTransformer bring also more understanding in GUI layout constraints. This work also introduce new loss functions and evaluation metrics specific to this task of GUI layout generation. Future work is to generate GUI from the layouts to complete the GUI design pipeline.





\printbibliography[heading=subbibintoc]

\end{document}